\documentclass[letterpaper]{article} 
\usepackage{aaai2026}  
\usepackage{times}  
\usepackage{helvet}  
\usepackage{courier}  
\usepackage[hyphens]{url}  
\usepackage{graphicx} 
\usepackage{natbib}  
\usepackage{caption} 
\usepackage{algorithm}
\usepackage{algorithmic}

\usepackage{epsfig}
\usepackage{amsmath}
\usepackage{amssymb}
\usepackage{booktabs}
\usepackage{multirow}
\usepackage{pifont}
\usepackage{bm}

\usepackage{colortbl}
\usepackage{xcolor}
\definecolor{mypink}{rgb}{.99,.91,.95}
\definecolor{mygray}{gray}{.9}

\usepackage{makecell}

\usepackage{newfloat}
\usepackage{listings}
\DeclareCaptionStyle{ruled}{labelfont=normalfont,labelsep=colon,strut=off} 
\floatstyle{ruled}
\newfloat{listing}{tb}{lst}{}
\floatname{listing}{Listing}
\title{Aggregating Diverse Cue Experts for AI-Generated Image Detection}
\author{
    Lei Tan\textsuperscript{\rm 1}, Shuwei Li\textsuperscript{\rm 1}, Mohan Kankanhalli\textsuperscript{\rm 1}, Robby T. Tan\textsuperscript{\rm 1, 2}
}
\affiliations{
    \textsuperscript{\rm 1}National University of Singapore,\\
    \textsuperscript{\rm 2}ASUS Intelligent Cloud Services (AICS)\\
    lei.tan@nus.edu.sg, shuwei@u.nus.edu, mohan@comp.nus.edu.sg, robby.tan@nus.edu.sg
}

\usepackage{bibentry}

\begin{document}

\maketitle

\begin{abstract}
	The rapid emergence of image synthesis models poses challenges to the generalization of AI-generated image detectors.
	However, existing methods often rely on model-specific features, leading to overfitting and poor generalization.
	In this paper, we introduce the Multi-Cue Aggregation Network (MCAN), a novel framework that integrates different yet complementary cues in a unified network. 
	MCAN employs a mixture-of-encoders adapter to dynamically process these cues, enabling more adaptive and robust feature representation.
	Our cues include the input image itself, which represents the overall content, and high-frequency components that emphasize edge details.
	Additionally, we introduce a Chromatic Inconsistency (CI) cue, which normalizes intensity values and captures noise information introduced during the image acquisition process in real images, making these noise patterns more distinguishable from those in AI-generated content.
	Unlike prior methods, MCAN's novelty lies in its unified multi-cue aggregation framework, which integrates spatial, frequency-domain, and chromaticity-based information for enhanced representation learning.
	These cues are intrinsically more indicative of real images, enhancing cross-model generalization.
	Extensive experiments on the \textbf{GenImage}, \textbf{Chameleon}, and \textbf{UniversalFakeDetect} benchmark validate the state-of-the-art performance of MCAN. In the GenImage dataset, MCAN outperforms the best state-of-the-art method by up to \textbf{7.4\%} in average \textbf{ACC} across eight different image generators.
\end{abstract}

\begin{figure}[t]
	\centering
	\includegraphics[width=0.95\columnwidth]{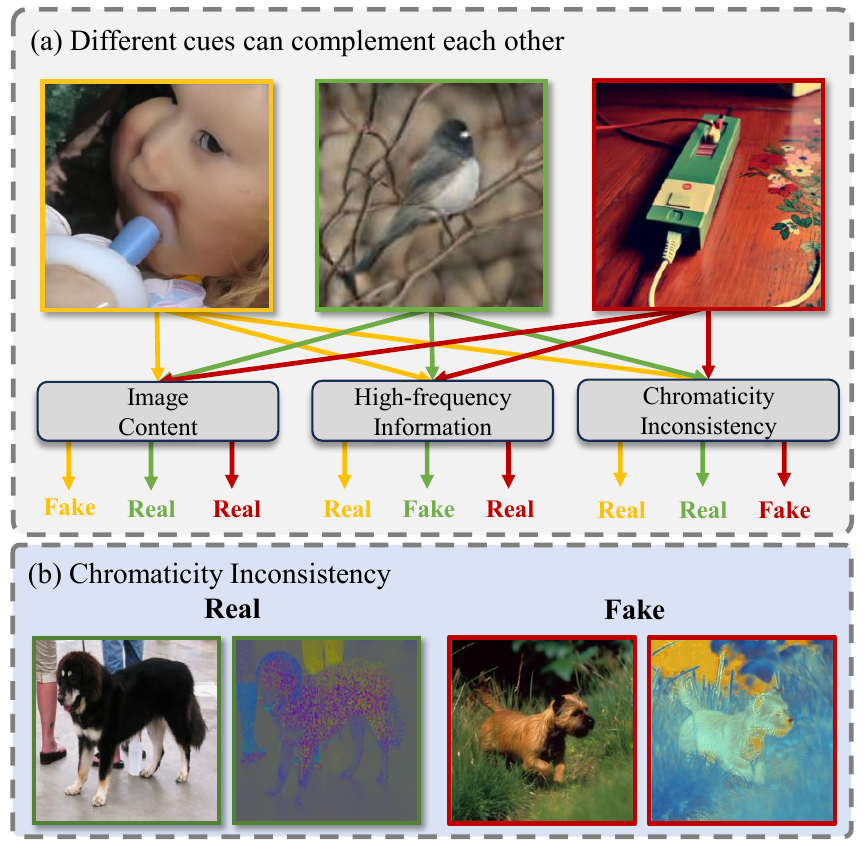}
	\caption{(a) Motivation for MCAN: In the top panel, yellow, green, and red represent the three generated images; Using multiple cues: image, frequency, and chromaticity enhances AI-generated image detection by leveraging the cues' complementary strengths. (b) Motivation for Chromaticity Inconsistency: In the bottom panel, real images show chromaticity inconsistencies due to noise, while fake images appear smoother with uniform chromaticity. 
	}
	\label{fig:motivation}
\end{figure}

\section{Introduction}
Distinguishing synthetic images from real ones is becoming increasingly challenging due to the rapid advancements in generative models. 
As new architectures emerge, they produce highly realistic images with fewer detectable artifacts, making traditional detection methods less effective. 
While many methods have been proposed (e.g.,~\cite{tan2024rethinking,sarkar2024shadows,cozzolino2025zero,wang2023dire,luo2024lare,ricker2024aeroblade}), most rely on model-specific features, leading to overfitting and poor generalization. 

A common approach in existing methods is to enhance the distinction between real and generated images using reconstruction error, which quantifies the discrepancy between an input image and its reconstructed version produced by a diffusion model~\cite{wang2023dire,luo2024lare,ricker2024aeroblade}.
While effective for many generative models, reconstruction error is highly dependent on the specific model used for reconstruction, limiting its generalizability.
Another approach involves leveraging high-frequency information~\cite{liu2024forgery,tan2024frequency}, which has demonstrated broader applicability across both GANs and diffusion models. However, high-frequency representations discard most semantic information, leading to failure cases where the lack of contextual details results in incorrect classifications.

The rise of large-scale pre-trained models has greatly advanced AI-generated image detection. UniFD~\cite{ojha2023towards} pioneered using image features from a frozen CLIP model to train a classifier, achieving strong detection performance. RBAD~\cite{cozzolino2024raising} demonstrated that CLIP can reliably detect images, even with limited generated samples. Fatformer~\cite{liu2024forgery} enhanced detection by fine-tuning CLIP with frequency-sensitive adapters. However, these methods rely on a single feature type, limiting their generalization ability.

To address the problem, as illustrated in Figure~\ref{fig:motivation}(a), we analyze three AI-generated images across classifiers, each using a different cue, and observe that these cues exhibit complementary properties in detecting synthetic content. For instance, AI-generated images missed by high-frequency representations often appear simple from an image content perspective.
Hence, in this paper, we propose integrating multiple cues into a unified network. Inspired by multi-modal learning, we treat these cues as multi-modal inputs and introduce the Multi-Cue Aggregation Network (MCAN). MCAN employs a Mixture-of-Encoder Adapter (MoEA), incorporating the mixture-of-experts concept to dynamically adapt diverse cues within a unified feature space.
Specifically, MoEA adopts a dynamic strategy that allows each image token to flexibly integrate different adapter encoders while utilizing a shared adapter decoder to project the fused cue representations.

Beyond the widely used original images, which primarily capture content, and high-frequency representations, which emphasize edge details, we introduce Chromaticity Inconsistency (CI), a novel representation designed to highlight noise differences between real and AI-generated images.
Noise components are challenging to extract directly from pixel differences due to variations in lighting intensity. To mitigate this, we apply a chromaticity transformation to minimize the influence of illumination. Theoretically, chromaticity should remain consistent across surfaces with uniform material and color temperature. However, real-world factors such as camera noise introduce inconsistencies at the pixel level in real images.
As shown in Figure~\ref{fig:motivation}(b), AI-generated images often appear smoother than real images due to the absence of such noise. Leveraging this distinction, we integrate CI with original and high-frequency representations to enhance the effectiveness of our MCAN.

To sum up, our key contributions are as follows: 
\begin{itemize} 
	\item \textbf{Chromaticity Inconsistency (CI)}: A new representation that mitigates lighting intensity effects through chromaticity-based transformation, highlighting noise differences between real and generated images.  
    \item \textbf{MCAN}: A novel framework that enhances AI-generated image detection by dynamically integrating spatial, frequency, and chromaticity through a unified multi-cue aggregation strategy, setting it apart from existing methods. 
	\item \textbf{Extensive Experiments}: In-depth evaluation on the GenImage, Chameleon, and UniversalFakeDetect benchmark shows MCAN superior performance. In the GenImage dataset, MCAN  outperforms the state-of-the-art method by up to 10.6\% in average Accuracy.
\end{itemize}

\section{Related Work}
\label{sec:related}

AI-generated image detection has long been a focus in the computer vision community, with increasing attention as generative models advance. Early approaches relied on hand-crafted cues such as chromatic aberration~\cite{mayer2018accurate}, color~\cite{mccloskey2019detecting}, saturation~\cite{mccloskey2019detecting}, blending~\cite{li2020face}, and reflections~\cite{o2012exposing}. These methods struggle with generalization as generation techniques evolve. Leveraging deep learning, neural networks have been applied to detect AI-generated images~\cite{liu2020global, wang2020cnnspot, marra2018detection}. 
CNNSpot~\cite{wang2020cnnspot} uses a standard CNN-based image classifier to achieve strong performance in detecting GAN-generated images. FreDect~\cite{frank2020leveraging} reveals that GAN images contain unique artifacts identifiable in the frequency domain and enhances detection by leveraging frequency-based representations.
MoE-FFD~\cite{kong2025moe} is an SOTA face forgery detection method that substantially improves model generalizability and reduces the parameter overhead.
PIM~\cite{kong2025pixel} effectively extracts inherent pixel-inconsistency forgery fingerprints and achieves SOTA performance in both generalization and robustness.
With the rise of diffusion models, detecting diffusion-generated images has gained traction. Ojha et al.~\cite{ojha2023towards} introduced a universal fake image detector using pre-trained vision-language models.
Fatformer~\cite{liu2024forgery} improves detection by fine-tuning CLIP with frequency-sensitive adapters and language-guided training. %
DIRE~\cite{wang2023dire} observed that diffusion-generated images are more easily reconstructed by diffusion models than real images. Lare$^2$~\cite{luo2024lare} improved DIRE in both efficiency and effectiveness. NPR~\cite{tan2024rethinking} refined upsampling in generative models, amplifying artifacts for better detection. ZED~\cite{cozzolino2025zero} identified discrepancies between real and generated images through lossless encoder coding differences, capturing them via a multi-resolution structure.
While these methods show promise by targeting specific features, the rapid emergence of diverse generative networks continues to challenge their generalizability.

\section{Proposed Method}
\begin{figure*}[t]
	\centering
	\includegraphics[width=2.0\columnwidth]{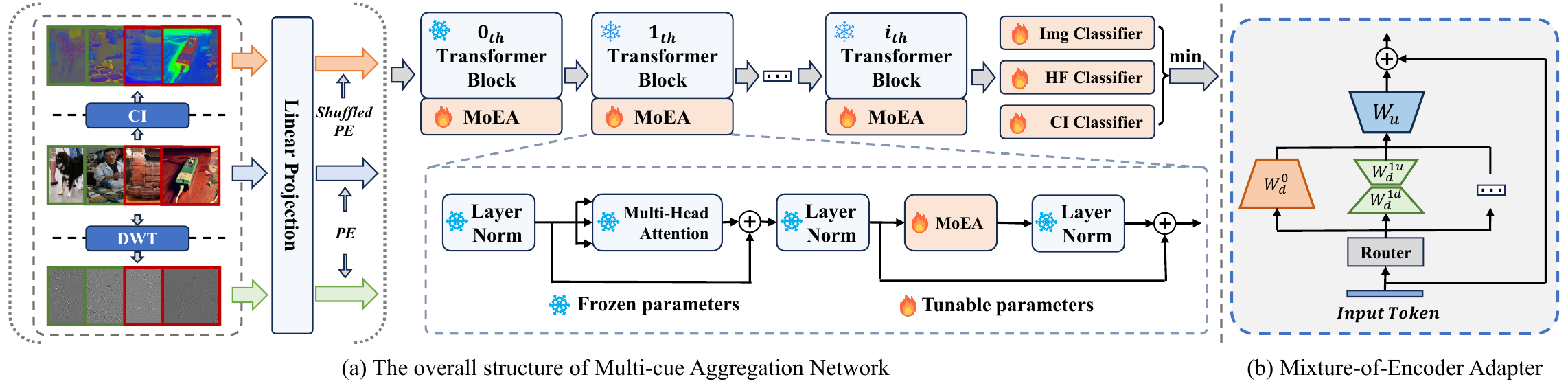}
	\caption{Overall architecture: MCAN combines image representation, high-frequency representation, and the novel chromaticity inconsistency as three distinct cues. To effectively integrate these cues, MCAN uses a mixture of encoder adapters that adapt efficiently to each cue's representation.}
	\label{fig:overall}
\end{figure*}
Figure~\ref{fig:overall} illustrates the architecture of our proposed Multi-Cues Aggregating Network (MCAN).
MCAN extends beyond image and high-frequency information~\cite{tan2024frequency,liu2024forgery} by incorporating chromaticity inconsistency as an additional cue.
It first applies a Discrete Wavelet Transform (DWT) to extract high-frequency features and uses a chromaticity inconsistency transformation to highlight noise discrepancies. A pre-trained CLIP ViT-B/16 model, with frozen parameters, serves as the backbone, while the Mixture-of-Encoder Adapter (MOEA) enables fine-tuning.
To enhance detection, separate classifiers process multiple cues, and the final classification is determined by aggregating their minimal outputs. 

\subsection{Chromaticity Inconsistency}
Noise introduced during image acquisition has been explored for detecting image manipulation and synthesis~\cite{mahdian2009using,lyu2014exposing,zhou2018learning}. However, due to the inherently low magnitude of noise signals, generative models struggle to capture them using conventional losses, like $L_1$ or $L_2$ loss. A key obstacle is variations in lighting intensity, which often dominate local discrepancies, making it difficult to identify noise from the original image. To address this, we leverage chromaticity information instead of raw images, reducing the impact of light intensity.

Following the common image formation model for Lambertian surfaces (e.g.~\cite{finlayson2005removal}), we express the response of each pixel $(x,y)$ in the camera sensor under a light source  across different wavelengths $\lambda$ as:
\begin{equation}
	\label{eq:lamb}
	{\rho_j} (x,y) = \sigma (x,y)\int_{\lambda_j}  {{E_j}(\lambda ,x,y)} S(\lambda ,x,y){Q}(\lambda )d\lambda,
\end{equation}
where $\lambda$ represents the wavelength, and $E(\lambda)$ and $S(\lambda)$ denote the spectral power distribution of the incident light and the surface spectral reflectance, respectively. ${Q}(\lambda)$ corresponds to the spectral sensitivity of the camera sensor. The subscript $j \in \{r,g,b\}$ indicates the channel in the spectral domain. $\sigma (x,y)$ represents the Lambertian reflection term, computed as the dot product between the surface normal and the illumination direction.

Following the Eq.~\eqref{eq:lamb}, we leverage Wien’s approximation to Planck’s law~\cite{wyszecki2000color} to parameterize the illuminant SPD by its light source color temperature $T$:
\begin{equation}
	\label{eq:illu}
	{E}(\lambda,T) = Ic_1\lambda^{-5}e^{\frac{-c_2}{T\lambda}},
\end{equation}
where $c_1$ and $c_2$ are constants, and $I$ is a variable controlling the overall intensity of the light. With Eq.~\eqref{eq:illu}, the Eq.~\eqref{eq:lamb} can be rewrite as: 
\begin{equation}
	\begin{split}
		\label{eq:lamb2}
		&{\rho_j} (x,y) = c_1\sigma (x,y)I(x,y)\int_{\lambda_j} {\lambda^{-5}e^{\frac{-c_2}{T(x,y)\lambda}}} F(\lambda ,x,y) d\lambda,
	\end{split}
\end{equation}
with $F(\lambda ,x,y) = S(\lambda ,x,y){Q}(\lambda ).$

We can then eliminate the light intensity $I$ by the chromaticity definition as:
\begin{equation}
	\frac{{{\rho _j}(x,y)}}{{{\rho _k}(x,y)}} = \frac{\int_{\lambda_j} {\lambda^{-5}e^{\frac{-c_2}{T(x,y)\lambda}}} S(\lambda ,x,y){Q_j}(\lambda )d\lambda}{{\int_{\lambda_k} {\lambda^{-5}e^{\frac{-c_2}{T(x,y)\lambda}}} S(\lambda ,x,y){Q_k}(\lambda )d\lambda}},
\end{equation}

\begin{figure}[t]
	\centering
    \includegraphics[width=1.0\columnwidth]{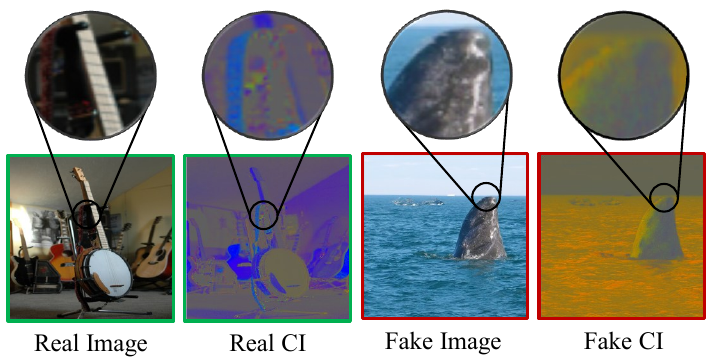}
  \caption{Visualization of Chromaticity Inconsistency (CI): Real images, affected by noise, show weaker consistency in CI images compared to generated images.}
  \label{fig:ci_vis}
\end{figure}
where $j$ and $k$ refer to two different color channels of the image. 

Since $Q(\lambda)$ and $S(\lambda)$ are intrinsic functions defined by the spectral sensitivity of the camera sensor and the surface material's reflectance, respectively, we make a mild assumption that local regions share uniform illumination conditions. This implies that the color temperature $T$ remains consistent within these regions, allowing the chromaticity information to be treated as constant locally. Consequently, for pixels corresponding to surfaces of the same material, denoted as $(x_a, y_a)$ and $(x_b, y_b)$, this relationship can be expressed as:
\begin{equation}
	\label{eq:ideal}
	\frac{{{\rho _j}(x_a, y_a)}}{{{\rho _k}(x_a,y_a)}} = \frac{{{\rho _j}(x_b, y_b)}}{{{\rho _k}(x_b,y_b)}}.
\end{equation}

Eq.~\eqref{eq:ideal} defines the ideal pixel relationship in a real image. However, when accounting for noise introduced during the acquisition process, the formulation becomes:
\begin{equation}
	\label{eq:noise}
	\frac{{{\rho _j}(x_a, y_a)}+{N _j}(x_a, y_a)}{{{\rho _k}(x_a,y_a)+{N _k}(x_a, y_a)}} \ne \frac{{{\rho _j}(x_b, y_b)+{N _j}(x_b, y_b)}}{{{\rho _k}(x_b,y_b)+{N _k}(x_b, y_b)}},
\end{equation}
where, $N_j$ and $N_k$ represent the random noise components in different color channels. 
From Eq.~\eqref{eq:noise}, the noise component can be extracted by computing the difference between pixels on the same surface. Leveraging this insight, we define the Chromaticity Inconsistency (CI) transformation as:
\begin{equation}
	\label{eq:CI}
	I_{ci} = [e^{-\frac{\rho _r}{\rho _g}},  e^{-\frac{\rho _g}{\rho _b}}, e^{-\frac{\rho _b}{\rho _r}}].
\end{equation}
To assess whether Eq.\eqref{eq:CI} effectively highlights noise differences between real and generated images, we visualize samples from the GenImage~\cite{zhu2023genimage} dataset in Figure~\ref{fig:ci_vis}. Due to the inherent noise introduced during image acquisition, real images exhibit lower consistency on the same material surface in CI images compared to generated ones.

\subsection{Multi-cue Aggregation Network}
Existing AI-generated image detection methods primarily focus on specific visual cues, such as image representations~\cite{ojha2023towards}, reconstruction errors~\cite{wang2023dire}, and others~\cite{sarkar2024shadows}. While some approaches integrate multiple cues~\cite{liu2024forgery,luo2024lare}, they often assign unequal importance to each, leading to suboptimal optimization and failing to fully leverage their complementary properties. To overcome this limitation, we take a multi-modal approach, incorporating image representation, high-frequency representation, and our proposed chromaticity inconsistency~\cite{liu2024forgery,luo2024lare} to build a robust detection model.

A crucial yet often overlooked factor is the difference in domain shifts between real and generated images across datasets. Real images exhibit relatively minor quality variations, whereas generated images undergo more significant shifts due to discrepancies in the training data and the structure of generative models. This discrepancy increases the risk of false positives, where generated images are misclassified as real.
To improve generalization, we aggregate multiple complementary cues. To this end, we introduce the Multi-Cue Aggregation Network (MCAN), which treats all cues as multi-modal inputs and encodes them within a unified framework. The following sections provide a detailed overview of MCAN’s architectural design.

\noindent\textbf{Position Embedding Shuffle} 
As shown in Figure~\ref{fig:ci_vis}, while CI amplifies noise, it also preserves some original image content, allowing the network to extract content information to some extent. Since MCAN already leverages original images for learning representations, we aim to minimize the influence of image content in CI representations.

In Vision Transformers (ViTs), positional embeddings determine the spatial locations of tokens. By perturbing the order of these embeddings, we disrupt the spatial structure of CI images, reducing content integrity while preserving noise characteristics. Specifically, for CI images, the shuffled positional embeddings are defined as:
\begin{equation}
	\label{eq:shuffle}
	p_{ci} = [p_\textrm{cls};\mathop{\underline{p_{s1};p_{s2};\dots;p_\textrm{sL}}}_\textrm{Shuffle Term}],
\end{equation}
where $L$ is the length of image tokens, $p_{si}$ refers to the shuffled position embedding in $i_{th}$ position.

\noindent\textbf{Mixture-of-encoder Adapter} 
In MCAN, different cues serve as distinct modalities. Extracting features through a shared network across these diverse modalities can lead to suboptimal performance~\cite{gou2023mixture}. To address this, we adopt the mixture-of-experts paradigm~\cite{riquelme2021scaling,chi2022representation} and introduce the Mixture-of-Encoder Adapter (MoEA) to effectively capture and integrate the unique characteristics of each cue. This design enhances feature discrimination and representation, ultimately improving detection performance.

Specifically, as illustrated in Figure~\ref{fig:overall} (b), given an input token $z_i\in \mathbb{R}^d$ and a set of $N$ encoder experts, we use a cosine-based router to normalize the token feature, ensuring stable routing scores. The routing function is defined as:
\begin{equation}
	g(z_i)=\sigma(\frac{W_1z_iW_e}{\tau \left \| W_1z_i\right \| \left \| W_e \right \|} ),
\end{equation}
where $\sigma(\cdot)$ denotes the softmax function, $\left \| \cdot  \right \| $ represents the $L_2$ normalization, and $\tau$ is the learnable temperature scalar. $W_1 \in \mathbb{R}^{d\times d_{e}}$ is the weight matrix of a linear projection, which maps the input token into a lower-dimensional subspace, while $W_e \in \mathbb{R}^{d_{e}\times N}$ serves as the expert embedding, transforming the feature into the final scoring distribution.

Leveraging the additive properties of different encoders, we integrate them through the scoring distribution, enabling the construction of a mixed encoder without separately computing results for each encoder. This significantly reduces computational cost during MoEA's inference stage. Consequently, the weight matrix of the mixed encoder, denoted as $W_d$, which encodes the input token into a $c$-dimensional subspace, is given by:
\begin{eqnarray}
	\label{eq:encoder}  \nonumber
	&W_d = (g(z_i)_{0}W_d^0 + g(z_i)_{1}W_d^{1} + \dots + g(z_i)_{N}W_d^N),\\ 
	& \text{ if  i $>$ 0}:W_d^i = W_d^{id}W_d^{iu},
\end{eqnarray}
where, $W_d^{0} \in \mathbb{R}^{d\times c}$, $W_d^{id} \in \mathbb{R}^{d\times \frac{c}{i}}$, and $W_d^{iu} \in \mathbb{R}^{\frac{c}{i}\times c}$.

Using an identical structure across all experts can lead to homogenized representations, limiting the diversity of learned features, especially when faced with heterogeneous inputs. To address this, we introduce $W_d^{id}$ and $W_d^{iu}$ to enhance expert diversity, ensuring that each expert contributes unique information to the model. This design increases the variability of expert outputs, leading to more effective feature extraction. Furthermore, since this transformation remains re-parameterizable, only a single mixed encoder needs to be maintained during inference, significantly reducing computational overhead. The overall processing of MoEA is formulated as:
\begin{equation}
	M(z_i) = z_iW_uW_d + z_i.
\end{equation}

\begin{table*}[t]
	\centering
    \renewcommand{\arraystretch}{1.1}
	\resizebox{175mm}{!}{
		\begin{tabular}{l|l|cccccccc|c}
			\toprule[1pt]
			\multirow{2}{*}{Method} & \multirow{2}{*}{Venue} & \multicolumn{8}{c|}{Testing Subset}                                                                                           & \multirow{2}{*}{\begin{tabular}[c]{@{}c@{}}Avg\\ Accuracy(\%)\end{tabular}} \\ \cline{3-10}
			& & Midjourney    & SDV1.4        & SDV1.5        & ADM           & GLIDE         & Wukong        & VQDM          & BigGAN        &                                                                         \\ \hline
            ResNet-50~\cite{he2016deepresnet}  &  CVPR'16   & 54.9 & 99.9 & 99.7 & 53.5 & 61.9 & 98.2 & 56.6 & 52.0 & 72.1 \\
            DeiT-S~\cite{touvron2021training}& ICML'21 & 55.6 & 99.9 & 99.8 & 49.8 & 58.1 & 98.9 & 56.9 & 53.5 & 71.6 \\
            Swin-T~\cite{liu2021swin}    &  ICCV'21   & 62.1 & 99.9 & 99.8 & 49.8 & 67.6 & 99.1 & 62.3 & 57.6 & 74.8 \\
            CNNSpot~\cite{wang2020cnnspot}    &  CVPR'20  & 52.8 & 96.3 & 95.9 & 50.1 & 39.8 & 78.6 & 53.4 & 46.8 & 64.2 \\
            Spec~\cite{zhang2019detecting} & WIFS'19  & 52.0 & 99.4 & 99.2 & 49.7 & 49.8 & 94.8 & 55.6 & 49.8 & 68.8 \\
            F3Net~\cite{qian2020thinking}  & ECCV'20  & 50.1 & 99.9 & 99.9 & 49.9 & 50.0 & 99.9 & 49.9 & 49.9 & 68.7 \\
            GramNet~\cite{liu2020global}   & CVPR'20  & 54.2 & 99.2 & 99.1 & 50.3 & 54.6 & 98.9 & 50.8 & 51.7 & 69.9 \\

			DIRE~\cite{wang2023dire}         & ICCV'23              & 65.8          & 99.7          & 99.7          & 54.5          & 58.1          & 99.4          & 54.3          & 49.8          & 72.7                                                                      \\
			LaRE$^{2}$~\cite{luo2024lare}   & CVPR'24              & 74.0          & \textbf{100.0}         & \textbf{99.9}          & 61.7          & 88.5          & \textbf{100.0}          & \textbf{97.2}         & 68.7          & 86.2 \\
			LaRE$^{2}$-ViT\dag~\cite{luo2024lare}   & CVPR'24              & 72.5          & 76.1          & 76.0          & 67.9          & 80.6          & 73.7          & 69.0          & 70.5 & 73.3 \\
			FatFormer\dag~\cite{liu2024forgery}    & CVPR'24              & 93.1          & 97.0          & 97.2          & 82.0          & 95.0          & 95.8          & 88.8          & 49.9          & 87.4                                                                      \\ 
            NPR~\cite{tan2024rethinking}    & CVPR'24              & 81.0          & 98.2          & 97.9          & 76.9          & 89.8          & 96.9         & 84.1          & 84.2          & 88.6                                                                      \\
            DRCT~\cite{chen2024drct}  & ICML'24              & 91.5          & 95.0          & 94.4          & 79.4          & 89.2          & 94.7          & 90.0         & 81.7          & 89.5                                                                      \\ 
            VIB-Net~\cite{zhang2025towards}  & CVPR'25              & 88.1          & 99.6          & 99.2          & 73.9          & 74.3          & 98.3          & 89.4         & -          & 88.9                                                                      \\ 
            AIDE~\cite{yan2025sanity}    & ICLR'25              & 79.4          & 99.7          & 99.8          & 78.5          & 91.8          & 98.7          & 80.3          & 66.9          & 86.9                                                                      \\ \hline
			\rowcolor{gray!15}
			MCAN                            & -                 & \textbf{94.3}          & 98.8          & 98.5          & \textbf{90.2}          & \textbf{98.6}          & \textbf{98.8}          & \textbf{96.8}          & \textbf{98.8}          & \textbf{96.9}                                                                      \\ \bottomrule
	\end{tabular}}
    \caption{Comparisons on between the proposed MCAN and state-of-the-art methods on the GenImage testing set. The symbol \dag indicates our reproduction using the source code on a CLIP-ViT-B/16 version.}
	\label{table1}
\end{table*}

\begin{table*}[t]
	\centering
    \renewcommand{\arraystretch}{1.1}
\resizebox{140mm}{!}{
\begin{tabular}{c|ccccccccc|c}
\toprule[1pt]
\multirow{2}{*}{Training Setting} & \multicolumn{10}{c}{Methods} \\ \cline{2-11}
                  & CNNSpot    & FreDect       & Fusing       & LNP         & UniFD        & DIRE          & Patchcraft       & NPR   & AIDE & MCAN                                                                      \\ \hline
ProGAN & 56.94 & 55.62 & 56.98 & 57.11 & 57.22 & 58.19 & 53.76 & 57.29 & 58.37 & \textbf{60.81}\\ 
SDV1.4                  & 60.11 & 56.86 & 57.07 & 55.63 & 55.62 & 59.71 & 56.32 & 58.13 & 62.60 & \textbf{69.61} \\
\bottomrule
\end{tabular}}
\caption{Comparisons between the proposed MCAN and state-of-the-art methods on the Chameleon dataset.}
\label{tab:cml}
\end{table*}

\subsection{Optimization}
To optimize the training of MCAN, alongside the binary cross entropy losses $\mathcal{L}_\textrm{img}$, $\mathcal{L}_{ci}$, $\mathcal{L}_{hf}$ for the classification, we introduce two additional loss functions according to the~\cite{riquelme2021scaling,li2023sparse}: importance loss ($\mathcal{L}_\textrm{imp}$) and entropy loss ($\mathcal{L}_\textrm{ent}$) to ensure both balanced expert assignment and effective specialization in MoEA. $\mathcal{L}_\textrm{imp}$ promotes a balanced usage of different experts, while the $\mathcal{L}_{ent}$ encourages each token to select a specific encoder expert in each layer. Therefore, the overall loss can be given as:
\begin{equation}
	\mathcal{L} = \mathcal{L}_\textrm{img} + \mathcal{L}_{ci} + \mathcal{L}_{hf} + \mathcal{L}_\textrm{imp} + \mathcal{L}_\textrm{ent}.
\end{equation}

In summary, MCAN adapts to diverse cues, enhancing AI-generated image detection and generalization. Chromaticity Inconsistency (CI) highlights noise discrepancies between real and generated images, critically complementing both image and high-frequency features.

\section{Experimental Results}
\paragraph{Datasets and Evaluation Metrics}
We evaluate our method on three challenging benchmarks that comprehensively cover various real and synthetic image domains: GenImage~\cite{zhu2023genimage}, Chameleon~\cite{yan2025sanity}, and UniversalFakeDetect~\cite{ojha2023towards}.
For training and evaluation, following the official protocol and previous works~\cite{zhu2023genimage,luo2024lare,yan2025sanity,tan2024rethinking}, we use Accuracy (ACC) as the evaluation metric.

\paragraph{Implementation Details}
We implement our MCAN using PyTorch and conduct all experiments on an RTX H100 GPU. The mini-batch size is set to 64, with each batch containing an equal number of randomly sampled real and generated images. The learning rate is fixed at $1 \times 10^{-4}$ during the training. We employ the CLIP-based ViT-B/16~\cite{radford2021learningclip} as the backbone. All images, in both training and testing phases, are resized to $224 \times 224$ without applying additional data augmentation strategies.

\begin{table*}[t]
\renewcommand\arraystretch{1.15}
    \centering
\resizebox{\textwidth}{!}{
    \begin{tabular}{c|c|c|c|c|c|c|c|c|c|c|c|c|c|c|c|c|c|c|c|c}
    \toprule
      \multirow{3}*{Methods}  & \multicolumn{6}{c|}{GAN} &  \multirow{3}*{\makecell[c]{Deep\\fakes}} & \multicolumn{2}{c|}{Low level} & \multicolumn{2}{c|}{Perceptual loss} &  \multirow{3}*{Guided} & \multicolumn{3}{c|}{LDM} & \multicolumn{3}{c|}{Glide} &  \multirow{3}*{Dalle}  &  \multirow{3}*{mAcc}\\ 
      
       \cline{2-7} \cline{9-10}  \cline{11-12}  \cline{14-16}  \cline{17-19}  
       ~ & \makecell[c]{Pro-\\GAN} & \makecell[c]{Cycle-\\GAN} & \makecell[c]{Big-\\GAN} & \makecell[c]{Style-\\GAN}  & \makecell[c]{Gau-\\GAN}  & \makecell[c]{Star-\\GAN} & ~ & \makecell[c]{SITD}& \makecell[c]{SAN}& \makecell[c]{CRN}& \makecell[c]{IMLE}& ~ & {\makecell[c]{200\\steps}}& {\makecell[c]{200\\w/cfg}}& {\makecell[c]{100\\steps}}& {\makecell[c]{100\\27}} & {\makecell[c]{50\\27}} & \makecell[c]{100\\10} & ~ & ~\\
       \hline
        CNNSpot & 100.0 & 85.2 & 70.2 & 85.7 & 79.0 & 91.7 & 53.5 & 66.7 & 48.7 & 86.3 & 86.3 & 60.1 & 54.0 & 55.0 & 54.1 & 60.8 & 63.8 & 65.7 & 55.6 & 69.6 \\
        Patchfor & 75.0 & 69.0 & 68.5 & 79.2 & 64.2 & 63.9 & 75.5 & 75.1 & 75.3 & 72.3 & 55.3 & 67.4 & 76.5 & 76.1 & 75.8 & 74.8 & 73.3 & 68.5 & 67.9 & 71.2 \\
        F3Net & 99.4 & 76.4 & 65.3 & 92.6 & 58.1 & 100.0 & 63.5 & 54.2 & 47.3 & 51.5 & 51.5 & 69.2 & 68.2 & 75.4 & 68.8 & 81.7 & 83.3 & 83.1 & 66.3 & 71.3  \\
        UniFD & 100.0 & 98.5 & 94.5 & 82.0 & 99.5 & 97.0 & 66.6 & 63.0 & 57.5 & 59.5 & 72.0 & 70.0 & 94.2 & 73.8 & 94.4 & 79.1 & 79.9 & 78.1 & 86.8 & 81.4 \\
        LGrad & 99.8 & 85.4 & 82.9 & 94.8 & 72.5 & 99.6 & 58.0 & 62.5 & 50.0 & 50.7 & 50.8 & 77.5 & 94.2 & 95.9 & 94.8 & 87.4 & 90.7 & 89.6 & 88.4 & 80.3  \\
        FreqNet & 97.9 & 95.8 & 90.5 & 97.6 & 90.2 & 93.4 & 97.4 & 88.9 & 59.0 & 71.9 & 67.4 & 86.7 & 84.6 & 99.6 & 65.6 & 85.7 & 97.4 & 88.2 & 59.1 & 85.1 \\
        NPR & 99.8 & 95.0 & 87.6 & 96.2 & 86.6 & 99.8 & 76.9 & 66.9 & 98.6 & 50.0 & 50.0 & 84.6 & 97.7 & 98.0 & 98.2 & 96.3 & 97.2 & 97.4 & 87.2 & 87.6 \\
        FatFormer & 99.9 & 99.3 & 99.5 & 97.2 & 99.4 & 99.8 & 93.2 & 81.1 & 68.0 & 69.5 & 69.5 & 76.0 & 98.6 & 94.9 & 98.7 & 94.4 & 94.7 & 94.2 & 98.8 & 90.9 \\
       \midrule
       \rowcolor{gray!15} MCAN & 100.0 & 99.6 & 98.8 & 97.0 & 99.3 & 100.0 & 94.0 & 86.7 & 68.9 & 87.3 & 87.3 & 70.9 & 98.8 & 94.2 & 98.5 & 97.4 & 97.2 & 97.1 & 98.8 & 93.3 \\
\bottomrule
    \end{tabular}
}
\caption{Comparisons between the proposed MCAN and state-of-the-art methods on the UniversalFakeDetect dataset.}
\label{tab:ud2}
\end{table*}

\begin{table*}[t]
	\centering
    \renewcommand{\arraystretch}{1.1}
	\resizebox{175mm}{!}{
		\begin{tabular}{l|cccccccc|c}
			\toprule[1pt]
			\multirow{2}{*}{Setting} & \multicolumn{8}{c|}{Testing Subset}                                                                                           & \multirow{2}{*}{\begin{tabular}[c]{@{}c@{}}Avg\\ ACC.(\%)\end{tabular}} \\ \cline{2-9}
			& Midjourney    & SDV1.4        & SDV1.5        & ADM           & GLIDE         & Wukong        & VQDM          & BigGAN        &                                                                         \\ \hline
			Img only   & 88.1          & \textbf{99.6}          & \textbf{99.5}          & 75.6          & 93.2          & 98.7          & 89.8          & 51.4          & 87.0                                                                    \\ 
			HF only  & 90.2          & 96.5          & 96.5          & 83.5          & 95.6          & 96.2          & 93.6          & 96.7          & 93.6                                                                    \\
			CI only & 87.5          & 96.9          & 96.7          & 81.7          & 85.2          & 96.7          & 94.1          & 51.7          & 86.3                                                                    \\
			CI-Shuffled Only & 89.1 & 97.2          & 97.0          & 86.7          & 86.5          & 96.9          & 95.0          & 70.0          & 89.8                                                                    \\
			Na\"ive Combination(Img, HF, CI-Shuffled) & 92.8          & 97.6          & 97.4          & 85.6          & 96.9          & 97.2          & 94.5          & 97.5          & 95.9                                                                      \\
			MCAN-(Img, HF)                            & 93.7          & 97.4          & 96.9          & 88.6          & 96.4          & 97.0          & 94.8          & 98.2          & 95.4  \\
			MCAN(Img, CI-Shuffled) & 93.3 & 98.6          & 98.5          & 88.8          & 97.5          & 98.4          & 96.5          & 69.5 & 92.6\\
			MCAN(HF, CI-Shuffled) & 94.1          & 98.2          & 98.3          & 89.8          & 95.5          & 97.3          & 96.3          & 98.8 & 96.0\\
			\rowcolor{gray!15}
			MCAN-(Img, HF, CI-Shuffled)                            & \textbf{94.3}          & 98.8          & 98.5          & \textbf{90.2}          & \textbf{98.6}          & \textbf{98.8}          & \textbf{96.8}          & \textbf{98.8}          & \textbf{96.9}                                                                      \\ \bottomrule
	\end{tabular}}
	\caption{Ablation study of the different components in MCAN: 'Img', 'CI', and 'HF' refer to the image cue, chromaticity inconsistency cue, and high-frequency cue, respectively.}
	\label{tab:ablation}
\end{table*}

\subsection{Comparison with State-of-the-Arts}
We compare the proposed MCAN with a broad set of state-of-the-art (SOTA) methods across three challenging benchmarks, including GenImage~\cite{zhu2023genimage}, Chameleon~\cite{yan2025sanity}, and UniversalFakeDetect~\cite{ojha2023towards}.

Table~\ref{table1} summarizes the average performance of different methods in the GenImage dataset. To ensure a fair comparison, we re-implement LaRE$^2$ with the same backbone as MCAN (CLIP ViT-B/16), as the original model uses CLIP-RN50. As shown in Table~\ref{table1}, MCAN achieves a leading average accuracy of 96.9\%, surpassing DRCT~\cite{chen2024drct} by a large margin of 7.4\%. Moreover, Table~\ref{table1} reveals that most existing methods suffer notable performance drops under cross-model protocols due to overfitting to single cues. In contrast, MCAN integrates multiple complementary visual cues, leading to consistently strong performance across all subsets and superior generalization.

We further evaluate MCAN on the challenging Chameleon dataset by comparing it against a comprehensive set of SOTA detectors, including CNNSpot~\cite{wang2020cnnspot}, FreDect~\cite{frank2020leveraging}, Fusing~\cite{ju2022fusing}, LNP~\cite{liu2022detecting},  UniFD~\cite{ojha2023towards}, DIRE~\cite{wang2023dire}, Patchcraft~\cite{zhong2023rich}, NPR~\cite{tan2024rethinking}, AIDE~\cite{yan2025sanity}. As shown in Table~\ref{tab:cml}, MCAN outperforms all SOTAs. On the ProGAN training protocol, it achieves 60.81\% accuracy, surpassing AIDE (58.37\%) by 2.44\%. The performance gap widens on the SDV1.4 training protocol, where MCAN reaches 69.61\%, outperforming AIDE (62.60\%) by 7.01\%. These results underscore MCAN’s capability in detecting high-quality synthetic content and its robustness across different generative paradigms.

On the UniversalFakeDetect dataset, MCAN is evaluated against methods specifically designed for generalization across multiple generative models. As shown in Tables~\ref{tab:ud2}, MCAN consistently outperforms prior approaches including CNNSpot~\cite{wang2020cnnspot}, Patchfor~\cite{chai2020makes}, F3Net~\cite{qian2020thinking}, UniFD~\cite{ojha2023towards}, LGrad~\cite{tan2023learning}, FreqNet~\cite{tan2024frequency}, NPR~\cite{tan2024rethinking}, Fatformer~\cite{liu2024forgery} in mean Accuracy (mAcc). Notably, MCAN improves upon UniFD~\cite{ojha2023towards}, which freezes the CLIP ViT-L/14 encoder and trains only a classifier on image features, by 11.9\% in mAcc. Compared to Fatformer~\cite{liu2024forgery}, which employs an adapter-based architecture, MCAN achieves 2.4\% higher mAcc. These findings further validate MCAN’s strong generalization capability and robustness to distribution shifts across generative techniques.

\begin{figure*}[t]
	\centering
	\includegraphics[width=2.1\columnwidth]{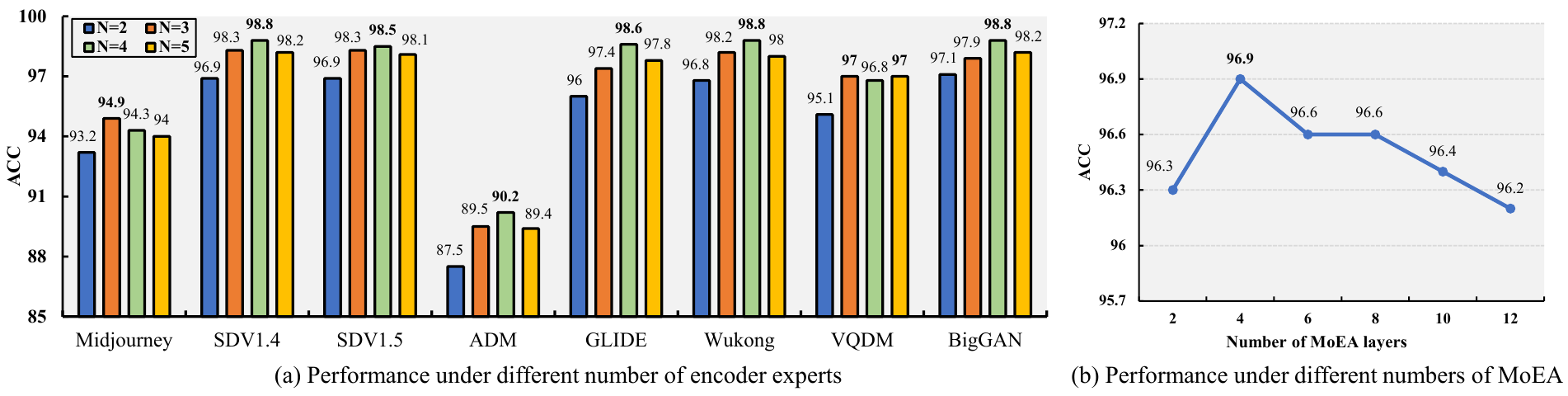}
	\caption{Performance of MCAN under structures: Optimal reaches when MoEA contains 4 experts in the last 4 blocks.}
	\label{fig:dn}
\end{figure*}

\begin{figure*}[t]
	\centering
	\includegraphics[width=2.1\columnwidth]{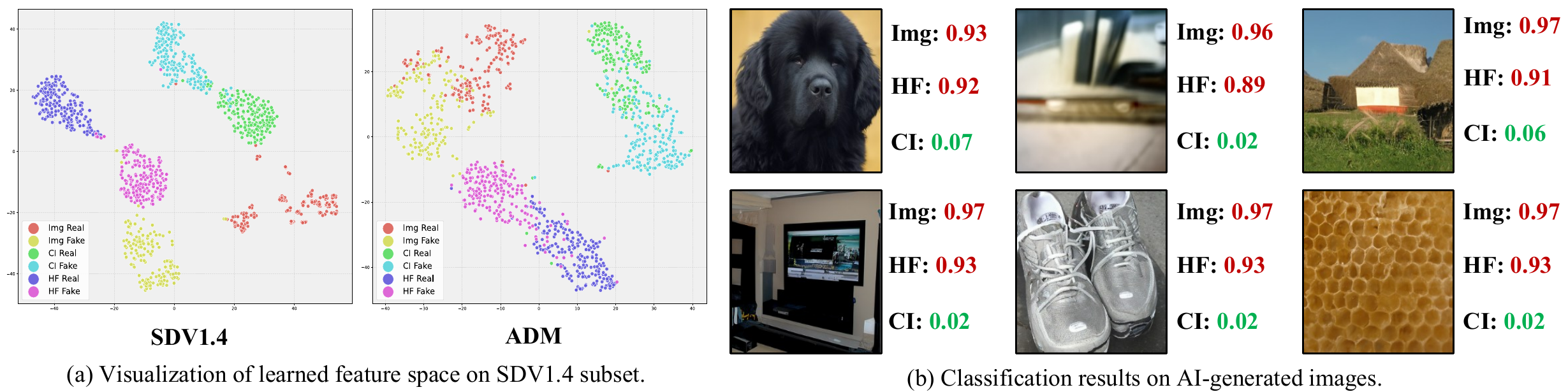}
	\caption{Visualization of learned feature space of MCAN (a) and classification results for generated images under different cues (b). (a) While all cues face generalization challenges, MCAN improves generalization by leveraging complementary features across cues. (b) To demonstrate the complementary nature of CI to 'Img' and 'HF', we specifically select failure cases for 'Img' and 'HF'. The values shown indicate the confidence scores from each cue, representing the likelihood that the corresponding image is classified as real.
	}
	\label{fig:feature_vis2}
\end{figure*}

\subsection{Ablation Studies}
we conduct comprehensive ablation studies on MCAN by incrementally integrating different modules into the baseline model to evaluate their contributions. Following LaRE$^{2}$\cite{luo2024lare}, we train the model on Stable Diffusion V1.4 and assess its performance across all eight subsets of GenImage. The results are summarized in Table~\ref{tab:ablation}.

The configurations in Table~\ref{tab:ablation} are defined as follows: '\textbf{Img only}' trains the baseline model solely on raw images, while '\textbf{HF only}' uses only high-frequency representations. '\textbf{CI only}' is trained exclusively on Chromaticity Inconsistency (CI) representations, whereas '\textbf{CI-shuffled}' applies a positional embedding shuffle strategy within CI training. '\textbf{Naïve Combination}' aggregates predictions from three independently trained models on these cues. Finally, '\textbf{MCAN}' represents our proposed model, which effectively integrates multiple cues using MoEAs.

From Table~\ref{tab:ablation}, incorporating positional embedding shuffling into CI training improves average accuracy by 3.5\% compared to the 'CI only' setting. This suggests that disrupting the spatial structure of CI images helps the model focus on noise patterns rather than image content, enhancing feature learning.

Since raw images, high-frequency components, and CI representations offer complementary information, simply aggregating predictions from separately trained networks already enhances performance. However, MCAN further optimizes this process by seamlessly integrating multiple cues within a unified network, achieving superior results.

To assess the impact of CI, we conducted an ablation study by combining CI with 'HF', 'Img', and 'HF+Img'. These combinations yielded performance improvements of 5.6\%, 2.4\%, and 1.5\%, respectively, highlighting the crucial role of CI in enhancing MCAN’s effectiveness.

\subsection{Discussion}

\paragraph{Number of Encoder Experts}  
In the MoE framework, the number of encoder experts plays a crucial role in model performance. To assess its impact, we vary the number of encoder experts from 2 to 5 and conduct a detailed performance comparison, as shown in Figure~\ref{fig:dn} (a). The results show that when the number of encoder experts is fewer than the number of cues, performance drops significantly, indicating that effective feature extraction requires sufficient expert capacity. However, increasing the number of encoder experts does not always yield further improvements. In most cases, the optimal performance is achieved with four encoder experts, reaching an average accuracy of 96.9\%.

\paragraph{Number of MoEA Layers}  
The placement of MoEA layers within the network is a critical design choice. To evaluate its impact, we conduct empirical experiments by inserting MoEA layers at different positions within the CLIP-B/16 model while keeping the remaining layers equipped with a single expert adapter following the MoEA structure. The results, shown in Figure~\ref{fig:dn} (b), provide key insights. Our findings reveal that integrating MoEA into every layer is unnecessary. In MCAN, the best performance is achieved when MoEA layers are inserted in the last four layers. This suggests that shallower layers primarily capture general features, while deeper layers focus on more specialized information. Based on this, MCAN integrates MoEA only in the last four layers of CLIP-B/16, optimizing both efficiency and effectiveness.

\paragraph{Visualization Results} 
To further demonstrate the effectiveness of MCAN, we train the model on the SDV1.4 subset and visualize the learned feature spaces on both SDV1.4 and the unseen ADM subset in Figure~\ref{fig:feature_vis2} (a). 
The results highlight a key challenge: generalization remains difficult when relying on individual cues alone. 
However, the inherent diversity among different cues enables each to provide a unique perspective for AI-generated image detection. 
In Figure~\ref{fig:feature_vis2} (b), we also visualize selected examples where 'Img' and 'HF' classifiers fail but 'CI' provides correct predictions. These images exhibit high visual quality and realistic content, leading to high-confidence predictions by 'Img' and 'HF' classifiers. However, subtle noise artifacts present in these images enable 'CI' to effectively identify them as fake.
This complementary nature helps mitigate generalization errors, enhancing MCAN’s robustness. These findings suggest that integrating multiple cues reduces the risk of overfitting associated with any single cue, ultimately improving both reliability and adaptability.

\section{Conclusion}
In this paper, we introduce the Multi-Cue Aggregation Network (MCAN), a novel framework for AI-generated image detection. MCAN effectively leverages the complementary properties of multiple cues and integrates them dynamically using a Mixture-of-Encoder Adapter (MoEA) within a unified network. In addition to the image and high-frequency cues, we introduce Chromaticity Inconsistency (CI), a new cue based on chromaticity that amplifies noise discrepancies between real and generated images. Given the complementary nature of these cues, integrating CI into MCAN further enhances detection performance. 

\bibliography{aaai2026}

@inproceedings{luo2024lare,
  title={LaRE\^{} 2: Latent Reconstruction Error Based Method for Diffusion-Generated Image Detection},
  author={Luo, Yunpeng and Du, Junlong and Yan, Ke and Ding, Shouhong},
  booktitle={Proceedings of the IEEE/CVF Conference on Computer Vision and Pattern Recognition},
  pages={17006--17015},
  year={2024}
}

@inproceedings{yan2025sanity,
  title={A sanity check for ai-generated image detection},
  author={Yan, Shilin and Li, Ouxiang and Cai, Jiayin and Hao, Yanbin and Jiang, Xiaolong and Hu, Yao and Xie, Weidi},
  booktitle={International Conference on Learning Representations},
  year={2025}
}

@inproceedings{liu2024forgery,
  title={Forgery-aware adaptive transformer for generalizable synthetic image detection},
  author={Liu, Huan and Tan, Zichang and Tan, Chuangchuang and Wei, Yunchao and Wang, Jingdong and Zhao, Yao},
  booktitle={Proceedings of the IEEE/CVF Conference on Computer Vision and Pattern Recognition},
  pages={10770--10780},
  year={2024}
}

@article{o2012exposing,
  title={Exposing photo manipulation with inconsistent reflections.},
  author={O'brien, James F and Farid, Hany},
  journal={ACM Trans. Graph.},
  volume={31},
  number={1},
  pages={4--1},
  year={2012}
}

@article{mayer2018accurate,
  title={Accurate and efficient image forgery detection using lateral chromatic aberration},
  author={Mayer, Owen and Stamm, Matthew C},
  journal={IEEE Transactions on information forensics and security},
  volume={13},
  number={7},
  pages={1762--1777},
  year={2018},
  publisher={IEEE}
}

@inproceedings{touvron2021training,
  title={Training data-efficient image transformers \& distillation through attention},
  author={Touvron, Hugo and Cord, Matthieu and Douze, Matthijs and Massa, Francisco and Sablayrolles, Alexandre and J{\'e}gou, Herv{\'e}},
  booktitle={International conference on machine learning},
  pages={10347--10357},
  year={2021},
  organization={PMLR}
}

@inproceedings{liu2021swin,
  title={Swin transformer: Hierarchical vision transformer using shifted windows},
  author={Liu, Ze and Lin, Yutong and Cao, Yue and Hu, Han and Wei, Yixuan and Zhang, Zheng and Lin, Stephen and Guo, Baining},
  booktitle={Proceedings of the IEEE/CVF international conference on computer vision},
  pages={10012--10022},
  year={2021}
}

@inproceedings{tan2024frequency,
  title={Frequency-Aware Deepfake Detection: Improving Generalizability through Frequency Space Domain Learning},
  author={Tan, Chuangchuang and Zhao, Yao and Wei, Shikui and Gu, Guanghua and Liu, Ping and Wei, Yunchao},
  booktitle={Proceedings of the AAAI Conference on Artificial Intelligence},
  volume={38},
  number={5},
  pages={5052--5060},
  year={2024}
}

@inproceedings{li2023sparse,
  title={Sparse Mixture-of-Experts are Domain Generalizable Learners},
  author={Li, Bo and Shen, Yifei and Yang, Jingkang and Wang, Yezhen and Ren, Jiawei and Che, Tong and Zhang, Jun and Liu, Ziwei},
  booktitle={International Conference on Learning Representations},
  year={2023}
}

@article{chi2022representation,
  title={On the representation collapse of sparse mixture of experts},
  author={Chi, Zewen and Dong, Li and Huang, Shaohan and Dai, Damai and Ma, Shuming and Patra, Barun and Singhal, Saksham and Bajaj, Payal and Song, Xia and Mao, Xian-Ling and others},
  journal={Advances in Neural Information Processing Systems},
  volume={35},
  pages={34600--34613},
  year={2022}
}

@article{riquelme2021scaling,
  title={Scaling vision with sparse mixture of experts},
  author={Riquelme, Carlos and Puigcerver, Joan and Mustafa, Basil and Neumann, Maxim and Jenatton, Rodolphe and Susano Pinto, Andr{\'e} and Keysers, Daniel and Houlsby, Neil},
  journal={Advances in Neural Information Processing Systems},
  volume={34},
  pages={8583--8595},
  year={2021}
}

@article{gou2023mixture,
  title={Mixture of cluster-conditional lora experts for vision-language instruction tuning},
  author={Gou, Yunhao and Liu, Zhili and Chen, Kai and Hong, Lanqing and Xu, Hang and Li, Aoxue and Yeung, Dit-Yan and Kwok, James T and Zhang, Yu},
  journal={arXiv preprint arXiv:2312.12379},
  year={2023}
}

@inproceedings{ojha2023towards,
  title={Towards universal fake image detectors that generalize across generative models},
  author={Ojha, Utkarsh and Li, Yuheng and Lee, Yong Jae},
  booktitle={Proceedings of the IEEE/CVF Conference on Computer Vision and Pattern Recognition},
  pages={24480--24489},
  year={2023}
}

@inproceedings{ricker2024aeroblade,
  title={AEROBLADE: Training-Free Detection of Latent Diffusion Images Using Autoencoder Reconstruction Error},
  author={Ricker, Jonas and Lukovnikov, Denis and Fischer, Asja},
  booktitle={Proceedings of the IEEE/CVF Conference on Computer Vision and Pattern Recognition},
  pages={9130--9140},
  year={2024}
}

@inproceedings{sarkar2024shadows,
  title={Shadows Don't Lie and Lines Can't Bend! Generative Models don't know Projective Geometry... for now},
  author={Sarkar, Ayush and Mai, Hanlin and Mahapatra, Amitabh and Lazebnik, Svetlana and Forsyth, David A and Bhattad, Anand},
  booktitle={Proceedings of the IEEE/CVF Conference on Computer Vision and Pattern Recognition},
  pages={28140--28149},
  year={2024}
}

@inproceedings{tan2023learning,
  title={Learning on gradients: Generalized artifacts representation for gan-generated images detection},
  author={Tan, Chuangchuang and Zhao, Yao and Wei, Shikui and Gu, Guanghua and Wei, Yunchao},
  booktitle={Proceedings of the IEEE/CVF Conference on Computer Vision and Pattern Recognition},
  pages={12105--12114},
  year={2023}
}

@inproceedings{tan2024rethinking,
  title={Rethinking the up-sampling operations in cnn-based generative network for generalizable deepfake detection},
  author={Tan, Chuangchuang and Zhao, Yao and Wei, Shikui and Gu, Guanghua and Liu, Ping and Wei, Yunchao},
  booktitle={Proceedings of the IEEE/CVF Conference on Computer Vision and Pattern Recognition},
  pages={28130--28139},
  year={2024}
}

@inproceedings{cozzolino2025zero,
  title={Zero-Shot Detection of AI-Generated Images},
  author={Cozzolino, Davide and Poggi, Giovanni and Nie{\ss}ner, Matthias and Verdoliva, Luisa},
  booktitle={European Conference on Computer Vision},
  pages={54--72},
  year={2025},
  organization={Springer}
}

@inproceedings{cozzolino2024raising,
  title={Raising the bar of ai-generated image detection with clip},
  author={Cozzolino, Davide and Poggi, Giovanni and Corvi, Riccardo and Nie{\ss}ner, Matthias and Verdoliva, Luisa},
  booktitle={Proceedings of the IEEE/CVF Conference on Computer Vision and Pattern Recognition},
  pages={4356--4366},
  year={2024}
}

@book{wyszecki2000color,
  title={Color science: concepts and methods, quantitative data and formulae},
  author={Wyszecki, G{\"u}nther and Stiles, Walter Stanley},
  volume={40},
  year={2000},
  publisher={John wiley \& sons}
}

@inproceedings{zhou2018learning,
  title={Learning rich features for image manipulation detection},
  author={Zhou, Peng and Han, Xintong and Morariu, Vlad I and Davis, Larry S},
  booktitle={Proceedings of the IEEE conference on computer vision and pattern recognition},
  pages={1053--1061},
  year={2018}
}

@article{mahdian2009using,
  title={Using noise inconsistencies for blind image forensics},
  author={Mahdian, Babak and Saic, Stanislav},
  journal={Image and vision computing},
  volume={27},
  number={10},
  pages={1497--1503},
  year={2009},
  publisher={Elsevier}
}

@article{lyu2014exposing,
  title={Exposing region splicing forgeries with blind local noise estimation},
  author={Lyu, Siwei and Pan, Xunyu and Zhang, Xing},
  journal={International journal of computer vision},
  volume={110},
  pages={202--221},
  year={2014},
  publisher={Springer}
}

@article{finlayson2005removal,
  title={On the removal of shadows from images},
  author={Finlayson, Graham D and Hordley, Steven D and Lu, Cheng and Drew, Mark S},
  journal={IEEE transactions on pattern analysis and machine intelligence},
  volume={28},
  number={1},
  pages={59--68},
  year={2005},
  publisher={IEEE}
}

@article{wang2023dire,
  title={DIRE for Diffusion-Generated Image Detection},
  author={Wang, Zhendong and Bao, Jianmin and Zhou, Wengang and Wang, Weilun and Hu, Hezhen and Chen, Hong and Li, Houqiang},
  journal={arXiv preprint arXiv:2303.09295},
  year={2023}
}

@article{kong2025moe,
  title={Moe-ffd: Mixture of experts for generalized and parameter-efficient face forgery detection},
  author={Kong, Chenqi and Luo, Anwei and Bao, Peijun and Yu, Yi and Li, Haoliang and Zheng, Zengwei and Wang, Shiqi and Kot, Alex C},
  journal={IEEE Transactions on Dependable and Secure Computing},
  year={2025},
  publisher={IEEE}
}

@article{kong2025pixel,
  title={Pixel-inconsistency modeling for image manipulation localization},
  author={Kong, Chenqi and Luo, Anwei and Wang, Shiqi and Li, Haoliang and Rocha, Anderson and Kot, Alex C},
  journal={IEEE Transactions on Pattern Analysis and Machine Intelligence},
  year={2025},
  publisher={IEEE}
}

@inproceedings{chai2020makes,
  title={What makes fake images detectable? understanding properties that generalize},
  author={Chai, Lucy and Bau, David and Lim, Ser-Nam and Isola, Phillip},
  booktitle={European Conference on Computer Vision},
  pages={103--120},
  year={2020},
  organization={Springer}
}

@article{zhong2023rich,
  title={Rich and poor texture contrast: A simple yet effective approach for ai-generated image detection},
  author={Zhong, Nan and Xu, Yiran and Qian, Zhenxing and Zhang, Xinpeng},
  journal={CoRR},
  year={2023}
}

@inproceedings{liu2022detecting,
  title={Detecting generated images by real images},
  author={Liu, Bo and Yang, Fan and Bi, Xiuli and Xiao, Bin and Li, Weisheng and Gao, Xinbo},
  booktitle={European Conference on Computer Vision},
  pages={95--110},
  year={2022},
  organization={Springer}
}

@inproceedings{ju2022fusing,
  title={Fusing global and local features for generalized ai-synthesized image detection},
  author={Ju, Yan and Jia, Shan and Ke, Lipeng and Xue, Hongfei and Nagano, Koki and Lyu, Siwei},
  booktitle={2022 IEEE International Conference on Image Processing (ICIP)},
  pages={3465--3469},
  year={2022},
  organization={IEEE}
}

@inproceedings{zhang2019detecting,
  title={Detecting and simulating artifacts in gan fake images},
  author={Zhang, Xu and Karaman, Svebor and Chang, Shih-Fu},
  booktitle={WIFS},
  pages={1--6},
  year={2019},
  organization={IEEE}
}

@inproceedings{frank2020leveraging,
  title={Leveraging frequency analysis for deep fake image recognition},
  author={Frank, Joel and Eisenhofer, Thorsten and Sch{\"o}nherr, Lea and Fischer, Asja and Kolossa, Dorothea and Holz, Thorsten},
  booktitle={ICML},
  pages={3247--3258},
  year={2020},
  organization={PMLR}
}

@inproceedings{mccloskey2019detecting,
  title={Detecting GAN-generated imagery using saturation cues},
  author={McCloskey, Scott and Albright, Michael},
  booktitle={ICIP},
  pages={4584--4588},
  year={2019},
  organization={IEEE}
}

@inproceedings{li2020face,
  title={Face x-ray for more general face forgery detection},
  author={Li, Lingzhi and Bao, Jianmin and Zhang, Ting and Yang, Hao and Chen, Dong and Wen, Fang and Guo, Baining},
  booktitle={CVPR},
  pages={5001--5010},
  year={2020}
}

@inproceedings{qian2020thinking,
  title={Thinking in frequency: Face forgery detection by mining frequency-aware clues},
  author={Qian, Yuyang and Yin, Guojun and Sheng, Lu and Chen, Zixuan and Shao, Jing},
  booktitle={ECCV},
  pages={86--103},
  year={2020},
  organization={Springer}
}

@inproceedings{liu2020global,
  title={Global texture enhancement for fake face detection in the wild},
  author={Liu, Zhengzhe and Qi, Xiaojuan and Torr, Philip HS},
  booktitle={CVPR},
  pages={8060--8069},
  year={2020}
}

@inproceedings{marra2018detection,
  title={Detection of gan-generated fake images over social networks},
  author={Marra, Francesco and Gragnaniello, Diego and Cozzolino, Davide and Verdoliva, Luisa},
  booktitle={MIPR},
  pages={384--389},
  year={2018},
  organization={IEEE}
}

@inproceedings{wang2020cnnspot,
  title={CNN-generated images are surprisingly easy to spot... for now},
  author={Wang, Sheng-Yu and Wang, Oliver and Zhang, Richard and Owens, Andrew and Efros, Alexei A},
  booktitle={CVPR},
  pages={8695--8704},
  year={2020}
}

@article{zhu2023genimage,
  title={Genimage: A million-scale benchmark for detecting ai-generated image},
  author={Zhu, Mingjian and Chen, Hanting and Yan, Qiangyu and Huang, Xudong and Lin, Guanyu and Li, Wei and Tu, Zhijun and Hu, Hailin and Hu, Jie and Wang, Yunhe},
  journal={Advances in Neural Information Processing Systems},
  volume={36},
  year={2024}
}

@inproceedings{radford2021learningclip,
  title={Learning transferable visual models from natural language supervision},
  author={Radford, Alec and Kim, Jong Wook and Hallacy, Chris and Ramesh, Aditya and Goh, Gabriel and Agarwal, Sandhini and Sastry, Girish and Askell, Amanda and Mishkin, Pamela and Clark, Jack and others},
  booktitle={ICML},
  pages={8748--8763},
  year={2021},
  organization={PMLR}
}

@inproceedings{he2016deepresnet,
  title={Deep residual learning for image recognition},
  author={He, Kaiming and Zhang, Xiangyu and Ren, Shaoqing and Sun, Jian},
  booktitle={CVPR},
  pages={770--778},
  year={2016}
}

@inproceedings{chen2024drct,
  title={Drct: Diffusion reconstruction contrastive training towards universal detection of diffusion generated images},
  author={Chen, Baoying and Zeng, Jishen and Yang, Jianquan and Yang, Rui},
  booktitle={Forty-first International Conference on Machine Learning},
  year={2024}
}

@inproceedings{zhang2025towards,
  title={Towards Universal AI-Generated Image Detection by Variational Information Bottleneck Network},
  author={Zhang, Haifeng and He, Qinghui and Bi, Xiuli and Li, Weisheng and Liu, Bo and Xiao, Bin},
  booktitle={Proceedings of the Computer Vision and Pattern Recognition Conference},
  pages={23828--23837},
  year={2025}
}

\end{document}